\documentclass[10pt,a4paper]{amsart}

\usepackage{times}
\usepackage[left=1.3in,right=1.3in,top=1.3in,bottom=1.3in]{geometry} 
\usepackage{amsmath}
\usepackage{enumitem}
\usepackage{color}
\usepackage[capitalise]{cleveref}

\usepackage{tikz,tikz-cd}

\usepackage{lipsum}
\usepackage{amsfonts}
\usepackage{graphicx}
\usepackage{epstopdf}
\usepackage{algorithmic}
\usepackage{caption}
\usepackage{subcaption}
\usepackage{bbm}

\newtheorem{lemma}{Lemma}
\newtheorem{proposition}[lemma]{Proposition}
\newtheorem{theorem}[lemma]{Theorem}
\numberwithin{lemma}{section}

\theoremstyle{definition}
\newtheorem{definition}[lemma]{Definition}

\theoremstyle{definition}
\newtheorem{remark}[lemma]{Remark}

\theoremstyle{definition}

\theoremstyle{definition}
\newtheorem{example}[lemma]{Example}

\theoremstyle{definition}

\graphicspath{{Pictures/}}

\numberwithin{equation}{section}

\newcommand{\R}{\mathbb{R}}

\newcommand{\Q}{\mathbb{Q}}

\newcommand{\B}{\mathcal{B}}
\newcommand{\F}{\mathcal{F}}
\newcommand{\M}{\mathcal{M}}
\newcommand{\Y}{\mathcal{Y}}
\newcommand{\eS}{\mathcal{S}}
\newcommand{\K}{\mathcal{K}}

\newcommand{\h}{h^p_{\bar{f}}}

\def\XXint#1#2#3{{\setbox0=\hbox{$#1{#2#3}{\int}$} 
\vcenter{\hbox{$#2#3$}}\kern-.5\wd0}}

\newcommand{\be}{\begin{equation}}
\newcommand{\en}{\end{equation}}

\DefineNamedColor{named}{JungleGreen} {cmyk}{0.99,0,0.52,0}

\usepackage{csquotes}

\title{Do stable neural networks exist for classification problems? -- A new view on stability in AI}

 \author{Z. N. D. Liu} 
\address{CCIMI, University of Cambridge}
\email{zndl2@cam.ac.uk} 
  
\author{A. C. Hansen} 
\address{DAMTP, University of Cambridge}
\email{a.hansen@damtp.cam.ac.uk}


\begin{document}

\maketitle

\begin{abstract}
In deep learning (DL) the instability phenomenon is widespread and well documented, most commonly using the classical measure of stability, the Lipschitz constant. While a small Lipchitz constant is traditionally viewed as guarantying stability,  it does not capture the instability phenomenon in DL for classification well. The reason is that a classification function -- which is the target function to be approximated -- is necessarily discontinuous, thus having an 'infinite' Lipchitz constant. 
As a result, the classical approach will deem every classification function unstable, yet basic classification functions a la 'is there a cat in the image?' will typically be locally very 'flat' -- and thus locally stable -- except at the decision boundary. The lack of an appropriate measure of stability hinders a rigorous theory for stability in DL, and consequently, there are no proper approximation theoretic results that can guarantee the existence of stable networks for classification functions.
In this paper we introduce a novel stability measure $\mathcal{S}(f)$, for any classification function $f$, appropriate to study the stability of discontinuous functions and their approximations. We further prove two approximation theorems:
First, for any $\epsilon > 0$ and any classification function $f$ on a \emph{compact set}, there is a neural network (NN) $\psi$, such that $\psi - f \neq 0$ only on a set of measure $< \epsilon$, moreover, $\mathcal{S}(\psi) \geq \mathcal{S}(f) - \epsilon$ (as accurate and stable as $f$ up to $\epsilon$). Second, for any classification function $f$ and $\epsilon > 0$, there exists a NN $\psi$ such that  $\psi = f$ on the set of points that are at least $\epsilon$ away from the decision boundary.

\end{abstract}

\section{Introduction}

With the advent of adversarial attacks in deep learning (DL) -- demonstrating universal instability of DL methods throughout the sciences \cite{akhtar2018threat, Maths_of_adv_attacs, carlini2018audio,Choi_IEEE,finlayson2019adversarial,madry1, fawzi4, fawzi1, fawzi3, Intriguing_properties,Tyukin_stealth} -- the necessity for the investigation of stability properties of neural networks (NN) became evident. The traditional approach of investigating the size of the Lipschitz constant is frequently adopted \cite{1Lip_loss,unser1,Certifiably_robust} with some quite remarkable results such as the work by Bubeck and Sellke \cite{bubeck2021a}, who proved a relation between the number of parameters and the Lipschitz constant of a NN. While the narrative of the Lipschitz constant is very useful for a wide range of scenarios, it should come as no surprise that it is unsuitable for describing discontinuous functions, which have  'infinite' Lipschitz constants.  As a consequence, the expectation of having an accurate NN (approximating the classification well) with a 'small' Lipschitz constant is unrealistic, as the target function to be approximated is necessarily unstable. This is particularly problematic for DL, as a major strength of DL lies in its application in image recognition, which is inherently a discontinuous task.  This is further emphasised by the empirical observations of instabilities and hallucinations in image recognition \cite{Hallucination_NatureM1,Anders2,Hallucination_NatureM2,Anders1,opt_big,Heaven_Nature,adcock3,raj2020improving,rodrigues1,madry2,l_inf,rethink}. The instability issue in DL is considered one of the key problems in modern AI research, as pointed out by Y. Bengio:
 "For the moment, however, no one has a fix on the overall problem of brittle AIs" (from 'Why deep-learning AIs are so easy to fool' \cite{Heaven_Nature}). This leads to the key problem addressed in this paper:
 \vspace{2mm}
 \begin{displayquote}
	\normalsize
	{\it Do stable neural networks exist for classification problems?}
\end{displayquote}
 \vspace{2mm}

Conceptually, there is a lack of a proper theory for the stability of discontinuous functions. Of course, one way would be to categorise all discontinuous functions as unstable, which while true, neglects the fact that there might be various degrees of instabilities for discontinuous functions. As an example a function such as the Heaviside step function seems intuitively more stable than for example the Dirichlet function, which is nowhere continuous. To tackle this issue we introduce a new stability measure, which we will call the \textit{class stability}, that is appropriate to study the stability of discontinuous functions and their approximations that captures this phenomenon through extending classical measure theory. The proposed stability measure focuses on the closest points with different functional values. This concept is also slowly being discovered and adapted by the machine learning community and is more commonly known as the `margin` \cite{Certifiably_robust}, which is a local measure of stability.  Our concept of the class stability extends this notion to the whole function on its entire domain, whilst also providing a way to compare the stability of different discontinuous functions. We also provide two different working definitions of the class stability, depending on whether the input space is discrete or continuous, where in the later the stability is defined in a measure theoretic way.

Finally, in the spirit of existing approximation papers \cite{AdcockHansenBook,SCI,devore2,adcock2,adcock1,Voigtlaender2,Celledoni,Anders3,dahl2011context,devore4,devore1,devore3,girshick2014rich, tyukin1,kutyniok2,bolcskei2,he2016deep,hinton2012deep,kutyniok1,bolcskei1,Voigtlaender1,pinkus_1999}, we prove the existence of NNs with class stabilities approximating the target function. Using results from approximation theory, analysis and measure theory, we prove two major theorems. The first one states that NNs are able to interpolate on sets that have a class stability of at least $\epsilon>0$, thereby proving that NNs can approximate any `stable' function (see \cref{remark:interpolation_thm}). The second is regarding the ability for NNs to approximate any function, such that the class stability of the NN is at most $\epsilon > 0$ smaller than the class stability of the target function. These results demonstrate that the class stability is appropriate to study stability for classification functions.

\section{Main result}

Our main contribution in this paper is the introduction of the `class stability' and two corresponding stability theorems for neural networks. The class stability is defined in \eqref{def:class_stability} in \cref{section:stability}. The intuitive notion behind it is that the class stability gives an average of distances to decision boundaries of the function.  The first of the two theorems deals with the restriction of classification functions to sets on which the classification functions have a class stability of at least $\epsilon>0$.  

\begin{theorem}[Interpolation theorem for stable sets]\label{interpolation_thm}
	 Let  $\M, \K \subset \R^d$, where $\K$ is compact, and $f:\M \rightarrow \Y \subset \mathbb{Z}^+$ be a non-constant classification function where $\Y$ is finite. Recall the extension $\overline{f}: \R^d \rightarrow \overline{\Y}$ defined in \eqref{def:extension}, where $\overline{\Y} = \Y \cup \{-1\}$. Define
	\begin{equation}
		\M_{\epsilon} := \{x \, \vert \, x \in \M,  \h(x) > \epsilon\}, \quad \epsilon > 0,
	\end{equation}
	as the $\epsilon$-stable set of $\overline{f}$, where 
	$\h$ is defined in \eqref{def:dist}.  
	Then, for any $\epsilon > 0$ and any continuous non-polynomial activation function $\rho$, which is continuously
differentiable at least at one point with nonzero derivative at that point, we have the following:
	\begin{enumerate}
		\item There exists a shallow neural network $\Psi_1 : \K \rightarrow \overline{\Y}$, with an activation function $\rho$, that interpolates $f$ on $\M_{\epsilon}$, in particular 
		      \begin{align}\label{result:shallow_network}
		      	p_{q}(\Psi_1(x)) = f(x) \quad \forall x \in \M_{\epsilon}\cap \K.  
		      \end{align}
		\item  There exists a neural network $\Psi_2:  \K \rightarrow \overline{\Y}$ with fixed width of $d+q+2$ and with an activation function $\rho$, that interpolates $f$ on $\M_{\epsilon}$, in particular
		      \begin{align}\label{result:deep_network}
		      	p_{q}(\Psi_2(x)) = f(x) \quad \forall x \in \M_{\epsilon}\cap \K. 
		      \end{align}
	\end{enumerate}
	Here $p_{q}$ is the class prediction function, given by \cref{def:classprediction}, that 'rounds' to discrete values, and $q = |\overline{\Y}|$.
\end{theorem}

\begin{remark}[Deep and Shallow neural networks]
A shallow network here means a neural network \cref{neural_network} with one layer, i.e. $L=1$, while the width of $d+q+2$ refers to $\max_{i=1, \hdots, L-1}{N_i} = d+q+2$.
\end{remark}

\begin{remark}[Interpretation of \cref{interpolation_thm}]\label{remark:interpolation_thm}
This theorem says that neural networks are able to interpolate any classification function restricted to compact sets on which the classification function attains some minimal class stability. In a simplified way, one can say that neural networks can interpolate on stable sets $M_\epsilon$, which are essentially the original set $M$ but with a small strip of width $\epsilon$ removed from the boundary of the set. This way we ensure that we are left with points that are at least $\epsilon$ away from the decision boundary, and then we simply interpolate on these sets. It is also important to mention that the approximation theorems utilised here do allow for arbitrary width in the shallow neural network case and for arbitrary depth in the deep neural network case. 
\end{remark}

The second theorem relates to the ability of neural networks to approximate the stability of the original classification function. The advantage of this theorem is that it also applies to the stability measure in a measure theoretic frameworks and is in a sense a generalisation of the first theorem.

\begin{theorem}[Universal stability approximation theorem for classification functions]\label{exist_stable}
	For any classification function $f: \M \subset \mathbb{R}^d \rightarrow \Y$,  where $\mathcal{M}$ is compact; any set $\{(x_i, f(x_i))\}_{i=1}^k$  such that $\h(x_i) > 0$ for all $i=1,\hdots,k$; and any $\epsilon_1, \,\epsilon_2 >0$, there exists a neural network $\psi \in \mathcal{NN}(\rho,n,m,1,\mathbb{N})$ such that the class stability of the neural network satisfies
	\begin{align}
		 \eS^p_\M(p_{q}(\psi)) \geq \eS^p_\M(\overline{f}) - \epsilon_1,
	\end{align}
	we can interpolate on the set
	\begin{align}
		p_{q}(\psi) = f(x_i) \quad i = 1,\hdots,k \, ,
	\end{align}
	and
	\begin{align}
		\mu(R) < \epsilon_2, \quad R := \{x \, \vert \, f(x) \neq p_{q}(\psi), x \in \M\}, 
	\end{align}
	where $\mu$ denotes the Lebesgue measure. 
\end{theorem}

\begin{remark}[Interpretation of \cref{exist_stable}]
This theorem proves that if one wants to use a neural network to approximate any fixed classification function, it is possible to achieve with a close to ideal stability, perfect precision (described by the second property) and an arbitrarily good accuracy (third property).
\end{remark}

\subsection{Computability and GHA vs existence of NNs -- Can the brittleness of AI be resolved?}
While our results produce a new framework for studying stability of NNs for classification problems, and provide theoretical guaranties for the existence of stable NNs for classification functions, the key issue of computability of such NNs  is left for future papers. Indeed, as demonstrated in \cite{Anders3}, based on the phenomenon of generalised hardness of approximation (GHA) \cite{opt_big}, there are many examples where one can prove the existence of NNs that can solve a  desired problem, but they cannot be computed beyond an approximation threshold $\epsilon_0 > 0$. Thus, what is needed is a theory that combines our existence theorems with GHA for which one can determine the approximation thresholds $\epsilon_0$ that will dictate the accuracy for which the NNs can be computed. This is related to the issue of NN dependency on the input.   

\begin{remark}[Non-compact domains and dependency on the inputs]
Note that our results demonstrate that on compact domains, one can always find a NN $\epsilon$-approximation $\psi$ to the desired classification function $f$, where the stability properties of $\psi$ are $\epsilon$ close to the stability properties of $f$. However, if the domain is not compact, this statement seizes to be true. The effect of this is that stable and accurate NN approximations to the classification function $f$ (on a non-compact domain) can still be found, however, the NN $\psi$ may have to depend on the input. Indeed, by choosing a compact domain $K_x$ based on the input $x$, one may use our theorem to find a NN $\psi_x$ such that $\psi_x(x) = f(x)$ and $\psi_x$ is stable on $K_x$. However, $\psi_x$ may have to change dimensions as a function of $x$. Moreover, if it is possible to make the mapping $x \mapsto \psi_x$ recursive is a big open problem. In particular, resolving the brittleness issue of moderns AI hinges on this question.  We mention in passing that there are papers in the machine learning community that deal with local decision boundary estimates in terms of certificates \cite{rethink}, that potentially provide a step towards computing class stable neural networks.
 \end{remark}

\subsection{Related work}

\begin{itemize}[leftmargin=12pt]

\item[] {\bf  \emph{Instability in AI:}} 
Our results are intimately linked to the instability phenomenon in AI methods -- which is widespread  -- and our results add theoretical understandings to this vast research program.  Notably, our work shares significant connections with the investigations conducted by F. Voigtlaender et al. \cite{Voigtlaender2}, which also deals with classification functions and their approximations via NNs. There has been significant work done on adversarial attacks by S. Moosavi-Dezfooli, A. Fawzi, and P. Frossard et al. \cite{fawzi1, fawzi4}. See also recent developments by D. Higham, I. Tyukin regarding vulnerabilities of neural networks et al. \cite{adversarial_ink, Tyukin_stealth}. Furthermore, our research aligns with the exploration of robust learning pursued by L. Bungert, G.Trillos et al. \cite{bungert2023trilos_binary} as well as by S. Wang, N. Si, J. Blanchet \cite{wang2023blanchet1}. The stability problem in neural network has also been extensively investigated by B. Adcock et al. \cite{Hallucination_NatureM1} and V. Antun et al. \cite{Anders3}.

\item[] {\bf  \emph{Existence vs computability of stable NNs:}}  There is a substantial literature on existence results of NNs \cite{yarotsky2018optimal, boelcskei2019optimal, Voigtlaender1}, see for example the aforementioned F. Voigtlaender et al. \cite{felix2}, review papers by A. Pinkus \cite{pinkus_1999} and the work by R. DeVore, B. Hanin, and G. Petrova \cite{devore3} and the references therein. Our work also utilises the approximation theorems obtained by P. Kidger and T. Lyons \cite{finite_width}. However, as established in \cite{Anders3} by M. Colbrook, V. Antun et al.\@, only a small subset of the NNs than can be proven to exist can be computed by algorithms. We also need to point out that following the framework of A. Chambolle and T. Pock \cite{Chambolle_Alg, Chambolle_Alg2}, the results in \cite{Anders3} demonstrate how -- under specific assumptions -- stable and accurate NNs can be computed. See also the work by P. Niyogi, S. Smale and S. Weinberger \cite{Smale_Weinberger} on existence results of algorithms for learning.

\end{itemize}

\section{Lipschitz constant and certificates}

In order to tackle the robustness of modern neural networks, researchers have applied various approaches. A standard way of looking at stability in general is to bound the Lipschitz constant of a neural network \cite{Intriguing_properties, Certifiably_robust, 1Lip_loss, l_inf}.  However, upon closer inspection, one would notice that any stability bound is almost always accompanied by a term that represents the `margin' of the neural network at the particular point. As an example take the Proposition 3.1. from \cite{rethink}. A local robustness certificate is given based on the quantity $\frac{c}{L}\text{margin}(f(x))$ and while this result is perfectly fine and true, it is essentially a lower bound on the local distance to the decision boundary.  In this section, we will demonstrate that the stability has very little to do with the actual Lipschitz constant, but rather the local distance to the decision boundary. In fact, there are several issues with the Lipschitz framework for stability. 

\subsection{Classification functions are inherently `unstable'}
The main issue is that classification functions are in general `unstable' in the sense of having an unbounded Lipschitz constant.  Intuitively, since classification functions are discrete functions, there has to be some region where the function looks like a step function which causes the Lipschitz constant to diverge. More generally, we can refer to the following proposition.

\begin{proposition}[Unbounded Lipschitz constant for classification functions] \label{prop:instability}
Let $\M$ be a connected subset of $\R^d$ and $f: \M \rightarrow \Y$ be a classification function that is not a constant function a.e. on  $\M$.  Then $f$ is not Lipschitz continuous.
\end{proposition}

The proof is elementary and simply follows from the fact that any non-constant discrete function on a connected domain has a discontinuity. 
Having this result, one should question the approach of trying to enforce a low Lipschitz constant for a neural network, where the target function has an unbounded Lipschitz constant. In the literature, there seems to be some acknowledgement to this fact, for instance in \cite{acc_vs_rob} demonstrates that the commonly used datasets have some minimal separation between different classes. In the context of our proposition, this is dropping the connectedness from our assumptions.  Furthermore, the issue of isolating the Lipschitz constant is highlighted by the fact that the classes themselves can be labeled by arbitrary numbers. This causes a problem for approaches such as the one in \cite{acc_vs_rob} where the distance between any two examples from different classes is assumed to be at least 2r, for some fixed value $r$ . As an example take the following functions

\begin{example}
	Let $H_1 : [-1,-\epsilon]\cup[\epsilon,1] \rightarrow \{0,1\}$ defined by
	\begin{align*}
		H_1(x) =
		\begin{cases}
		1 \quad &x>0, \\
		0 \quad &x<0.
		\end{cases}
	\end{align*}
	Similarly we define the function $H_2 : [-1,-\epsilon]\cup[\epsilon,1] \rightarrow \{0,1000\}$ defined by
	\begin{align*}
		H_2(x) =
		\begin{cases}
		1000 \quad &x>0, \\
		0 \quad &x<0.
		\end{cases}
	\end{align*}
	Clearly both of these functions could be used to describe the same classification problem by assigning $0$ to one class, $1$ to the other in the first function example or by assigning $0$ to one class, $1000$ to the other class in the second function example. Intuitively, the stability should not change as both functions are describing the same function, but the second function has a much higher Lipschitz constant. 
\end{example}

Of course, the standard way of encoding different classes is by doing a one-hot encoding, where different classes get assigned different unit basis vectors in $\R^n$ and therefore it might be a bit unnatural to talk about encoding the classes with non-unit vectors. There is however another catch, which is that one could rescale the inputs as well, to change the Lipschitz constant 

\begin{example}
	As before, we define a step function 	Let $H_3 : [-\frac{1}{1000},-\frac{\epsilon}{1000}]\cup[\frac{\epsilon}{1000},\frac{1}{1000}] \rightarrow \{0,1\}$ defined by
	\begin{align*}
		H_3(x) =
		\begin{cases}
		1 \quad &x>0, \\
		0 \quad &x<0.
		\end{cases}
	\end{align*}
	This function, unlike the function $H_2$ has a natural one-hot encoding, but differs in the 'transformation' we have applied to the input. As before, this function has a much higher Lipschitz constant, but the nature of the stability should not have really changed.
\end{example} 

In light of these examples, we propose a slightly different approach on how to understand the stability of classification functions, which are inherently `unstable' in the classical sense. We would like to emphasise that this does not necessarily contradict any existing work, in fact, we will demonstrate that it supports a lot of the existing research on robustness certificates. The aim of this work is to provide a framework in which functions which have so far been considered unstable, could be categorised into different types of stabilities.

\section{Examples of different stability boundaries}\label{example}
In this section we will give examples of functions that all have an unbounded Lipschitz constant, yet somehow one could consider them of different stability. These examples will also be used to demonstrate desired properties of a more general stability measure.

\begin{example}
	Let $ f_1, f_2, f_3 : [-1, 1] \rightarrow \{ -1,1 \} $ be defined by:
\begin{align*}
		f_1(x) = sgn(x),
\end{align*}
\begin{align*}
		f_2(x) =
		\begin{cases}
		-sgn(x) \quad & \text{if } x \in \{ -0.5, 0.5\}, \\
		sgn(x) \quad  & \text{otherwise},                
		\end{cases}
	\end{align*}
	and
\begin{align*}
		f_3(x) =
		\begin{cases}
		sgn(x) \quad  & \text{if } x \in \Q ,             \\
		-sgn(x) \quad & \text{if } x \in \R \setminus \Q. 
		\end{cases}
	\end{align*}
	Here the function $sgn : \R \rightarrow \{ -1, 1 \}$ is the sign function (for the sake of the argument we will assign 0 as positive),  i.e.
	\begin{align*}
		sgn(x) =                     
		\begin{cases}                
		1 \quad \text{if } x \geq 0, \\
		-1 \quad \text{if } x < 0.   
		\end{cases}                  
	\end{align*}
	
	\setlength{\tabcolsep}{2pt} 
	
	\begin{figure}[htbp]
	       \begin{subfigure}[t]{0.3\textwidth}
		  \centering
		 \includegraphics[width=\linewidth]{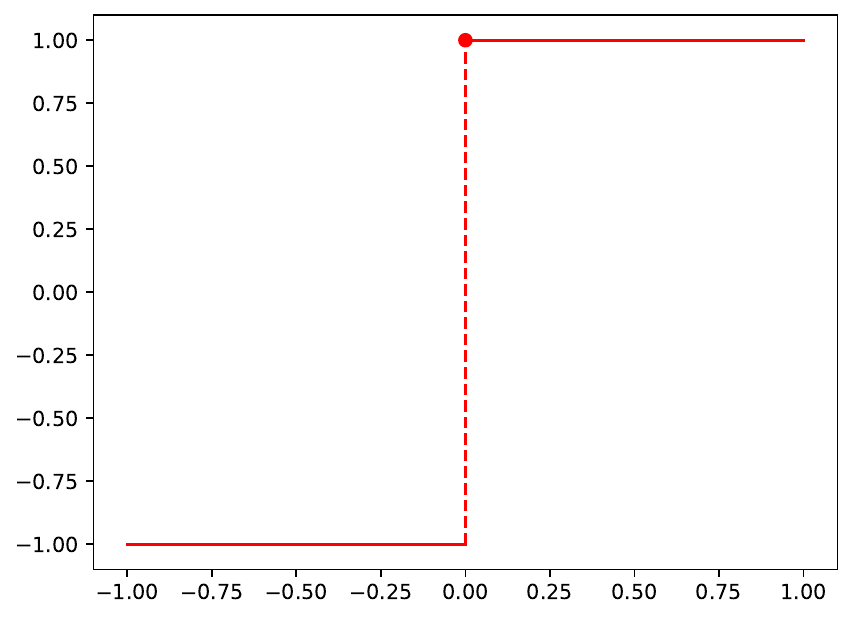}
		  \caption{Standard step function.}
	     \end{subfigure}
	     \hfill
	       \begin{subfigure}[t]{0.3\textwidth}
		  \centering
		  \includegraphics[width=\linewidth]{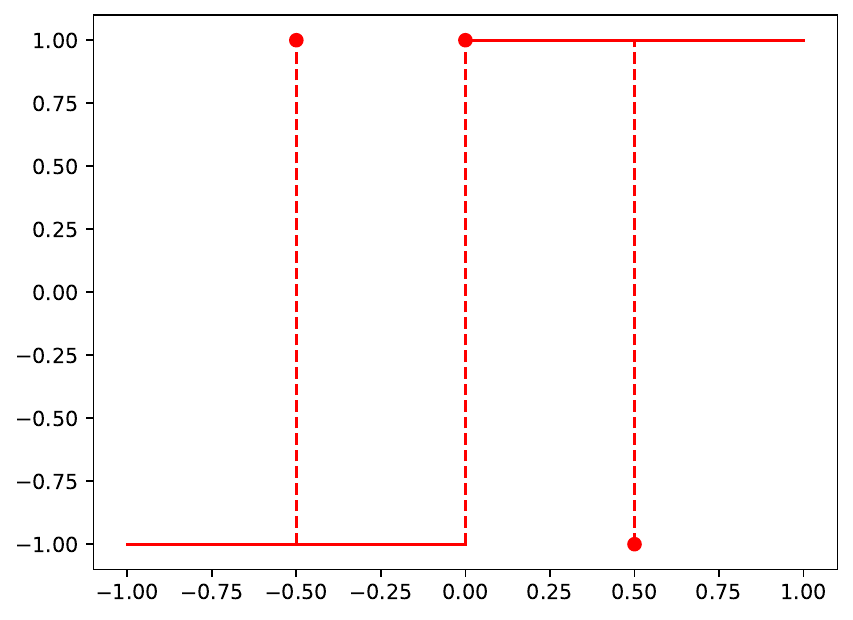}
		  \caption{Step function with extra discontinuities, making it more `unstable'.}
	     \end{subfigure}
	     \hfill
	       \begin{subfigure}[t]{0.3\textwidth}
		  \centering
		  \includegraphics[width=\linewidth]{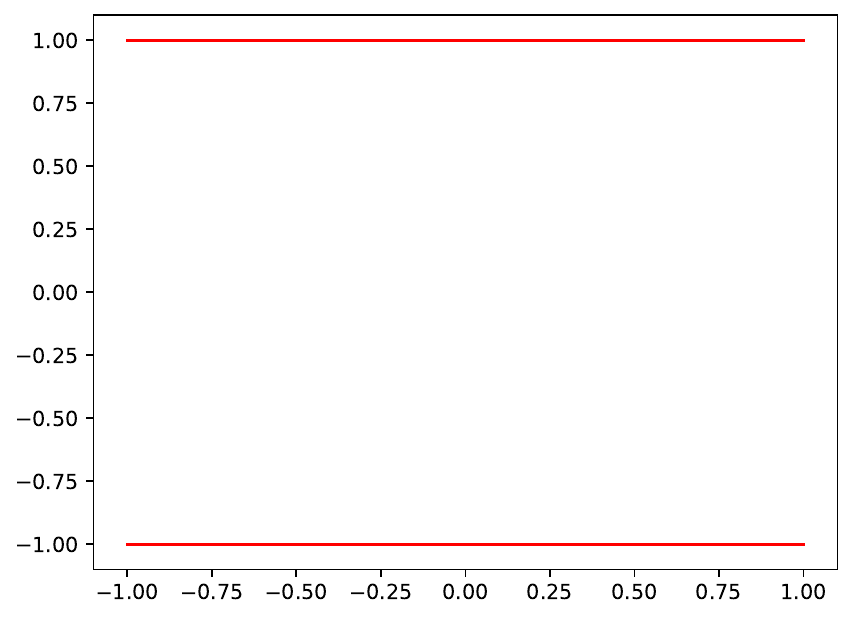}
		  \caption{A function that seems as unstable as it is possible.}
	     \end{subfigure}
	     \hfill
	        \caption{Different classes of unstable classification functions.}
	        \label{fig:functions}
	\end{figure}
	
	Let us briefly analyse these functions.  One could argue that the functions $f_1$, $f_2$, $f_3$ have
	different stability properties, most notably it is in the `count' of the discontinuities.  As illustrated in  \cref{fig:functions} we see that $f_2$ is just a more unstable version of $f_1$, with $f_3$ being a minefield of instabilities.
	This motivates us to define a local measure which takes into account the discontinuities but also the
	position of them, since a point close to the discontinuity would be more unstable in the sense of `What is the smallest perturbation needed to change the output of the function?'.
	
\end{example}

\section{Definitions}
Before we introduce the stability measure, we will have to define a few terms.
\begin{definition}[Classification Function]
Let $f: \M \rightarrow \Y$ be a function we are trying to learn where $\mathcal{M} \subset \R^d$ is the \textit{input domain} ($d$ is the dimension of the input) and $\Y \subset \mathbb{Z}^+$ a finite subset.
\end{definition}
\begin{definition}[Extension of a classification function] 
	We define the extension of the classification function $f : \M \rightarrow \Y$ to $\R^d$ as $\overline{f}: \R^d \rightarrow \overline{\Y}$ such that
	\begin{align}\label{def:extension}
		\overline{f}(x) =
		\begin{cases}
		f(x) \quad & \text{if } x \in \M , \\
		-1 \quad   & \text{otherwise} ,     
		\end{cases}
	\end{align}
	where $\overline{\Y} = \Y \cup \{-1\}$.
\end{definition}

\begin{definition}[Neural Network]\label{neural_network}
	Let $\mathcal{NN}^{\rho}_{\mathbf{N},L,d}$ where $\mathbf{N} = (N_L = |\Y|, N_{L-1},\hdots,N_1,N_0 = d)$ denote the set of all \textit{L}-layer neural networks. That is, all mappings $\phi : \R^d \rightarrow \R^{N_L}$ of the form:
	$$\phi(x) = W_L(\rho(W_{L-1}(\rho(\hdots\rho (W_1(x))\hdots)))), \quad x \in \R^d,$$
where $W_l : \R^{N_{l-1}} \rightarrow \R^{N_l}, 1 \leq l \leq L$ is an affine mapping and $\rho : \R \rightarrow \R$ is a function (called the activation function) which acts component-wise (Note that $W_L : \R^{N_{L-1}} \rightarrow \R^{|\Y|}$). Typically this function is given by $\rho (x) = \max\{0,x\}$.

\end{definition}

Throughout this paper we will also need to define specific sets of neural networks as they are crucial to approximation theorems. To this end we will use the following notation.

\begin{definition}[Class of Neural Networks]
Let $\mathcal{NN}(\rho, n,m, D,W)$ denote the set of neural networks $\mathcal{NN}^{\rho}_{\mathbf{N},L,d}$ with an activation function $\rho$,  input dimension $n$, output dimension $m$, depth $D$ and width $W$. In relation to the previous definition this means
\[
		\rho = \rho, \quad L = D, \quad d = n, \quad N_L = m, \quad \max_{i=1, \hdots, L-1}{N_i} = W.
		\]
We will also denote the neural network class with unbounded depth by $\mathcal{NN}(\rho, n,m, \mathbb{N},W)$, and similarly the neural network class with unbounded width by $\mathcal{NN}(\rho, n,m, D,\mathbb{N})$.
\end{definition}

\begin{definition}[Class Prediction Function]
	For a given $n \in \mathbb{Z^+}$ we define the \textit{class prediction function} $p_n:\R^n \rightarrow \{1, \hdots, n\}$ as 
	\begin{align} \label{def:classprediction}
		p_n(x) = \min\{i : x_i \geq x_j,  \forall j \in \{1, \hdots, n\} \} .
	\end{align}
\end{definition}

The class prediction function has the same function as the `argmax' function in for example the numpy library of python. This function takes a vector and returns the index of the element that has the highest value of all elements. If there are multiple such indices that satisfy the maximality, we return the first index.

By training a neural network on a classification task we mean that we want to approximate a classification function $f$, more precisely, its extension. To illustrate why we want the extension, imagine something simple as MNIST. We have 10 target classes, hence $\Y = \{1, 2, \dots, 10\}$ ('zero' is represented by 10 and each other number is represented by itself). Then we either want to learn $f$ which labels well-defined images correctly, while labelling undefined images randomly, or we want to learn $\overline{f}$ where we label undefined images as $-1$. Here $f$ is the ground truth (might be debatable whether it actually exists, but for the purpose of the argument assume it does).

\begin{definition}[Accuracy of a neural network]
	The accuracy of a neural network $\psi \in \mathcal{NN}^{\rho}_{\mathbf{N},L,d}$ on a set $\B \subset \M$ with respect to a classification function $f: \M \rightarrow \Y$ is defined as 
	\begin{align*}
		\mathbf{A}_{\B, f}(\psi) := \mu(\{x \, \vert \, \psi(x)=f(x), \, x\in \B\}).
	\end{align*}
	Here $\mu$ is the Lebesgue measure.
\end{definition}

\section{Alternative measure for 'robustness'}\label{section:stability}

As mentioned in the second section, we would like to define some finite stabilities to functions that can have an unbounded Lipschitz constant. Here we propose an alternative measure for stability for discrete functions. First of all we need to define what we mean by the distance to the decision boundary.

\begin{definition}[Distance to the decision boundary]
	For the extension of a classification function $\overline{f} : \R^d \rightarrow \overline{\Y}$ and a real number $1 \leq p \leq \infty$, we define $\h: \R^d \rightarrow \R^+$ the \textit{$L^p$-distance to the decision boundary} as
	\begin{align}\label{def:dist}
		\h(x) = \inf \{ \|x-z\|_p : \overline{f}(x) \neq \overline{f}(z), \, z \in \R^d   \}. 
	\end{align}
\end{definition}

It is easy to check that this definition indeed captures the intuitive notion of the `distance to the decision boundary', since the decision boundary is really just the closest place where the label flips.  Having the local stability measure, we can now
proceed to defining a global measure which would help us differentiate the different types of stabilities of for example functions $f_1$, $f_2$ and $f_3$. To assess the stability of a compact set $A \subset \R^d $, we define the stability of a function $\overline{f}$ to be the following:

\begin{definition}[Class stability of discrete function]
	Let $\overline{f} : \R^d \rightarrow \overline{\Y}$  be a extension of a classification function and $A \subset \R^d $ a compact set. Then for a real number $1 \leq p \leq \infty$, we define the \textit{$L_p$-stability of $\overline{f}$ on $A$} to be
	\begin{align*}
		\h(A) = \int_A \h \, d \mu .
	\end{align*}
	We call this stability measure the \textbf{class stability} of the function $\overline{f}$ on the set $A$.
\end{definition}
If the original classification function was defined on a compact set $\M \subset \R^d$ then we define the \textit{$L_p$-stability of $\overline{f}$} to be
\begin{align}\label{def:class_stability}
	\eS^p_{\M}(\overline{f}) = \int_{\M} \h \, d\mu ,
\end{align}
which we will reference as just the \textbf{class stability} of the function $\overline{f}$.

Let us now have a look at the $L_1$-stability of the functions $\overline{f_1}$, $\overline{f_2}$ and $\overline{f_3}$ on the compact set $\M = [-1,1]$.  For $f_1$ the distance to the decision boundary for a point $x$ simply
becomes $h^1_{\bar{f}}(x) = \|x\|$ and thus a simple calculation yields $\eS^1(\bar{f_1}) = 1$. Similarly we can compute the other values where we get $\eS^1(\bar{f_2}) = 0.5$ and $\eS^1(\bar{f_3}) = 0$.
The actual values are not that important as they do depend on the $L_p$ norm chosen, but what is convenient about this measure is that it does quantify $f_3$ as completely unstable.  In fact, in a way this function is chosen
to be one of the worst kinds as any perturbation anywhere may yield an extreme change.

\subsection{Properties of the class stability}
One interesting property of the class stability of a function is that it prefers uneven sets. What we mean by that is illustrated by the following example. Consider two classification functions $f_1, f_4: \M=[-1,1]\rightarrow \{ -1, 1\}$ where
\begin{align*}
	f_1(x)  = sgn(x),  \quad f_4(x)  = sgn\big(x+\frac{1}{2}\big). 
\end{align*}
The $L_1$ class stability of these functions on $\M$ are 1 and $\frac{5}{4}$ correspondingly. In fact, it is true for any $p>0$ that the $L_p$ norm of $f_1$ is lower than for $f_4$.
\begin{figure}[ht]
	\centering
	\begin{subfigure}[t]{0.45\textwidth}
	\centering
	\includegraphics[width=\linewidth]{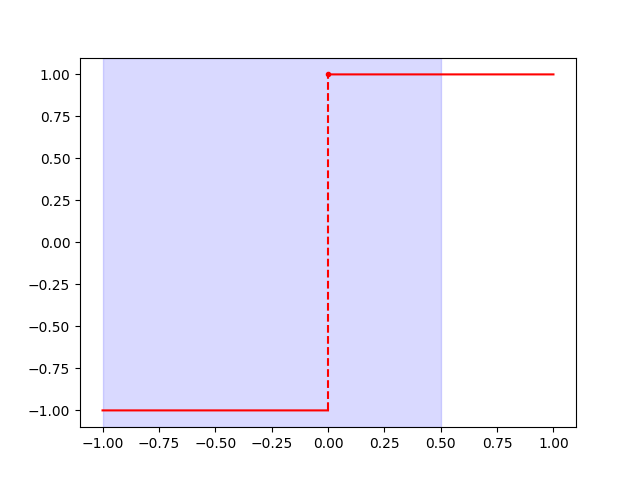}
	\caption{$f_1$, the original step function.}
	\end{subfigure}
	\begin{subfigure}[t]{0.45\textwidth}
	\centering
	\includegraphics[width=\linewidth]{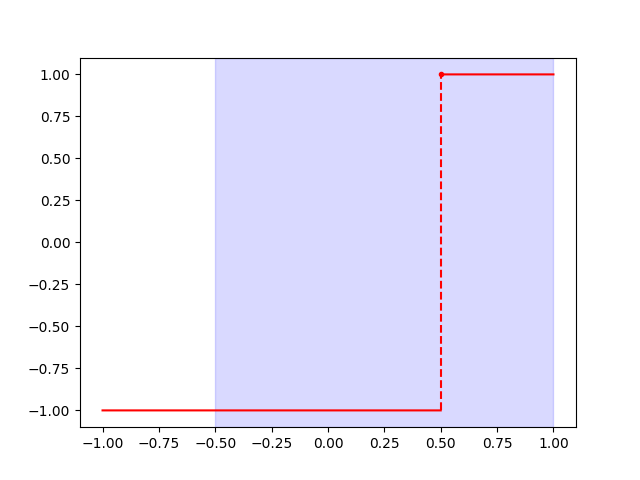}
	\caption{$f_4$, the shifted step function.}
	\end{subfigure}
	\caption{Step functions with differently placed steps.}
	\label{fig:different_step}
\end{figure}

We can see from \cref{fig:different_step} that in both functions, there is a region (shaded blue) for which the points have the exact same stability properties as the relative distance to the decision boundary remains the same. For the remaining points, we can see that the remaining portion $f_4$ is more stable than the remaining portion of $f_1$.  This property makes sense in the context of how the average stability of the function looks like. If the instability is hidden away from most points, then in some sense this is more beneficial to the overall stability.

\subsection{Class stability of specific sets}
In a lot of cases it is useful to be able to describe the class stability of simple sets. We will provide the derivation of the stabilities of a ball and cube in n dimensions. We will start with the simpler case which is the cube.  First we define the following two sets. 
\begin{align*}
	\mathfrak{S}^n_a = \{x \in \R^n : x \in [-a,a ]\},
\end{align*}
which is a cube centred at the origin with side lengths $2a$ and 
\begin{align*}
	\mathfrak{B}^n_a = \{x \in \R^n : |x|_2 \leq a\},
\end{align*}
the unit ball centred at the origin with radius $a$.

\begin{example}[Class stability of the cube]
	We first define the indicator function for the cube to be a classification function $f: \M \rightarrow \Y$ where $\M = \mathfrak{S}^n_a$ and 
	\begin{align*}
		f(x) = 1 \quad \forall x \in \M .
	\end{align*}
	Having defined the indicator function for the cube, we want to find the value of $\h(\M)$. To work this out we split the cube into symmetrical pieces in the following way
	\begin{align*}
		T_i= \{(x_1, x_2, \hdots, x_n) \in \mathfrak{S}^n_a : |x_j| \leq |x_i|, \, \forall j\neq i  \}.
	\end{align*}
	It is obvious to see that there are $n$ such sets in total and that their union gives us $\mathfrak{S}^n_a$.  One should note that the sets $T_i$ are not pairwise disjoint, however, their intersections have zero Lebesgue measure, so we can simply compute the class stability on these sets and their sum will give us the class stability of the whole set. The nice property about these sets is that for any point $x \in T_i$ we have $\h(x) = a - |x_i|$ as the decision boundary is the boundary of the set $[-a,a]$. Thus we simply calculate the class stability of the set $T_i$ as 
	\begin{align*}
		\h(T_i) = \int_{T_i} \h \, d\mu = \int_{-a}^{a}\int_{\mathfrak{S}^{n-1}_{|x_i|}}(a - |x_i|) \, d\mu dx_i  = \\
		2\int_{0}^{a}\int_{\mathfrak{S}^{n-1}_{|x_i|}}(a - x_i )\, d\mu dx_i = 2\int_{0}^{a}(2x_i)^{n-1}(a - x_i) \, dx_i = 
		2^n\frac{a^{n+1}}{n(n+1)}.
	\end{align*}
	Now since there are $n$ sets $T_i$ in $\mathfrak{S}^n_a$ we have the class stability of the whole set
	\begin{align*}
		\h(\mathfrak{S}^n_a) = 2^n\frac{a^{n+1}}{n+1}. 
	\end{align*}
	One interesting aspect here is that the class stability of a $n$-dimensional unit cube is independent of the norm chosen.  
\end{example}
\begin{example}[Class stability of the $L_2$ ball]
	In order to talk about the class stability in this case we need to choose a value for $p$. The natural choice for this in the case of a $L_2$ ball is $p = 2$. We can use the spherical symmetry to calculate the class stability. First of all, we recall that the expression for the surface of a $n$-dimensional ball with radius $r$ is given by
	\begin{align*}
		\frac{2\pi^{\frac{n}{2}}}{\Gamma(\frac{n}{2})} r^{(n-1)}.
	\end{align*}
	The class stability of the ball with radius $R$ then simply becomes
	\begin{align*}
		h^{2}_{\bar{f}}(\mathfrak{B}^n_R) = \int^{R}_{0}\frac{2\pi^{\frac{n}{2}}}{\Gamma(\frac{n}{2})} r^{(n-1)}(R-r)dr = \frac{2\pi^{\frac{n}{2}}}{\Gamma(\frac{n}{2})}\frac{R^{n+1}}{n(n+1)}.
	\end{align*}
\end{example}

The next natural question one might ask is whether the cube or the ball is more stable if they have the same volume (p =2). If we fix $a$ then the volume a $n$-dimensional cube $\mathfrak{S}^n_a$, then its volume is 
$
	V(\mathfrak{S}^n_a)=2^n a^n.
$
To match this we need the radius $R$ of the ball to satisfy
\begin{align*}
	\frac{\pi^{\frac{n}{2}}}{\Gamma(\frac{n}{2}+1)}R^n = 2^n a^n, 
	R = 2a\sqrt[n]{\frac{\Gamma(\frac{n}{2}+1)}{\pi^{\frac{n}{2}}}}.
\end{align*}
Thus we can compute the ratio of the class stabilities when the two objects have the same volume.
\begin{align*}
	\frac{h^{2}_{\bar{f}}(\mathfrak{B}^n_R)}{h^{2}_{\bar{f}}(\mathfrak{S}^n_a)} = \frac{2\pi^{\frac{n}{2}}2^{n+1} a^{n+1} \Gamma(\frac{n}{2}+1)^{\frac{n+1}{n}}}{\Gamma(\frac{n}{2})n(n+1)\pi^{\frac{n+1}{2}}} \frac{(n+1)}{2^n a^{n+1}}
	= \frac{4 \Gamma(\frac{n}{2}+1)^{\frac{n+1}{n}}}{\sqrt{\pi} \Gamma(\frac{n}{2})n} = 
	\frac{2\Gamma(\frac{n}{2}+1)^{\frac{1}{n}}}{\sqrt{\pi}} \rightarrow \infty,
\end{align*}
as $n \rightarrow \infty.$
The divergence comes from the application of Stirling's approximation.

\section{Proof of \cref{interpolation_thm}}

We are now equipped to prove our first main result. This concerns the existence of class stable neural networks. We will prove that for arbitrary depth (with fixed depth) or arbitrary width (with fixed depth) ReLu networks, stable neural networks exist for classification tasks that are in some sense well separated.  Our proof relies on the following two approximation results, the first being the classical approximation theorem for single layer neural networks.

\begin{theorem}[Universal approximation theorem \cite{pinkus_1999}]
Let $\rho \in C(\R)$ (continuous functions on $\R$) and assume $\rho$ is not a polynomial. Then $\mathcal{NN}(\rho,n,m,1,\mathbb{N})$ (the class of single layer neural networks with an activation function of $\rho$) is dense in $C(R)$.
\end{theorem}

The second theorem is a newer result that proves the universal approximation property for fixed width neural networks.

\begin{theorem}[P. Kidger,T. Lyons \cite{finite_width}]\label{thrm:Lyons}
	Let $\rho : \R \rightarrow \R$  be any non-affine continuous function which is continuously differentiable at at least one point, with nonzero
	 derivative at that point. Let $\K \subset \R$ be compact. Then $\mathcal{NN}(\rho,n,m,\mathbb{N},n+m+2)$(the class of neural networks with input dimension $n$, output dimension $m$ and width of at most $n+m+2$) is dense in $C(K; \R^m)$ with respect to the uniform norm.
	\end{theorem}

Before we prove \cref{interpolation_thm}, we will first prove a lemma. We start by defining the following functions. For each $i \in \overline{\Y}$ let us define the functions $H_i : \M \rightarrow \R$ as:
\begin{align}\label{eq:vector_distance}
	H_i(x) =
	\begin{cases}
	\h(x) \quad & \bar{f}(x) = i, \\
	0 \quad                & otherwise.      
	\end{cases}
\end{align}
This function can be thought of as an element-wise version of the distance to the decision boundary \cref{def:dist}.

\begin{lemma}\label{h_continuous}
	$H_i$ is continuous for all $i \in \overline{\Y}$
\end{lemma}
\begin{proof} Let $\{x_m\}$ be a sequence in $\K$ with $x_m \rightarrow x'$ as $m \rightarrow \infty$, where $x' \in \K$.   First we take care of the simple case where $\overline{f}(x') \neq i$. Then we know that $H_i(x') = 0$ and that for $x_m$ we have 
	$
		0 \leq H_i(x_m) \leq \|x_m -x'\|_p. 
	$
	Thus $H_i(x_m) \rightarrow H_i(x')$ as $m \rightarrow \infty$.
	Therefore we can assume $\overline{f}(x') = i$ in which case we distinguish three cases.
	
	\textit{Case 1 : $\exists j \in \mathbb{N}$ such that $\overline{f}(x_m) = i, \, \forall m>j$.} Pick an $\epsilon>0$. Then there exists a $l\in\mathbb{Z}$ such that
$
\|x_m - x'\|_{p} < \epsilon/2
$ for all $m>l.$ 
As $\overline{f}(x') = i$, it follows by the definition of $\h$, that there must exist a sequence of $\{z'_\alpha\}_{\alpha = 0}^{\infty}$ such that
	\begin{align*}
		\|x' - z'_\alpha\|_p \rightarrow \h(x') \quad \text{as } \alpha \rightarrow \infty, \text{ with } \overline{f}(z'_\alpha) \neq i.
	\end{align*}
	This also means that there exists a $\beta' \in \mathbb{Z}$ such that
	\[
	\|x' - z'_\alpha\|_p< \h(x') + \epsilon/2 \quad \forall \alpha > \beta', 
	\]
	hence
	\[
		\h(x_m) \leq \|x_m - z'_\alpha\|_{p} 
		\leq \|x_m - x'\|_{p} + \|x' - z'_\alpha\|_p <  \h(x') + \epsilon \quad \forall \alpha >\beta' \text{ and } m>l. 
	\]
	Notice that since $f(x_m) = i$ we also have a sequence $\{z_\alpha\}_{\alpha = 0}^{\infty}$ such that
	\begin{align*}
		\|x_m - z_\alpha\|_p \rightarrow \h(x_m) \quad \text{as }\alpha \rightarrow \infty,
	\end{align*}
	 $\forall m>l$.
	This also means that there exists a $\beta \in \mathbb{Z}$ such that
	\begin{align*}
		\|x_m - z_\alpha\|_p< \h(x_m) + \epsilon/2 \quad \forall \alpha > \beta\text{ and }m>l, 
	\end{align*}
	hence
	\begin{align*}
		\h(x') \leq \|x' - z_\alpha\|_{p} \leq \|x_m - x'\|_{p} + \|x_m - z_\alpha\|_p <  \h(x_m) + \epsilon \quad \forall \alpha >\beta\text{ and }m>l. 
	\end{align*}
	Putting these together we obtain $\|\h(x') - \h(x_m)\| < \epsilon \quad \forall m>l, \epsilon >0$. Thus $ \h(x_m) \rightarrow \h(x') $ as $m \rightarrow \infty$ and therefore $H_i(x_m) \rightarrow H_i(x')$ as $m \rightarrow \infty$.
	
\textit{Case 2: $\exists j \in \mathbb{N}$ such that $\overline{f}(x_m) \neq i, \, \forall m>j$.} In this case $\h(x') = 0$, since the subsequence has only points containing points that do not map to label $i$, whereas $\overline{f}(x') = i$. Similarly, $\|x_m - x'\|_p$ serves as an upper bound for $\h(x_m)$ for all $m>j$, but since $x_m \rightarrow x'$ as $m \rightarrow \infty$, we must also have $\h(x_m)\rightarrow \h(x')$.  

\textit{Case 3: $\forall j \in \mathbb{Z} \quad \exists m,l > j$ such that $\overline{f}(x_m) = i $ and $\overline{f}(x_l) \neq i$.}
In this case there exists a subsequence $\{x_{h_k}\}_{k=1}^\infty$ such that $\overline{f}(x_{h_k}) \neq i $ for all $ k \in \mathbb{Z}$ and $x_{h_k} \rightarrow x'$ as $k \rightarrow \infty$. This means that $\h(x') = 0$.  To show that $\h(x_m)\rightarrow 0$ as $m \rightarrow \infty$ we use the fact that the sequence is also a Cauchy sequence, and that elements that map to label $i$ and ones that do not map to label $i$ occur infinitely many times in the sequence.
	
Combining these gives us $H_i(x_m) \rightarrow H_i(x')$ as $m \rightarrow \infty$ as required.
\end{proof}

With this lemma we are now ready to prove our first main result \cref{interpolation_thm}.
\begin{proof}[Proof of Theorem \ref{interpolation_thm}]
	The proof will rely on two steps. First we show that we can find a continuous function $g : \K \rightarrow [0,1]^{q}$ that satisfies
	\begin{align*}
		p_{q}\circ g(x) = f(x)  \quad \forall x\in \M_{\epsilon}\cap \K .
	\end{align*}
	Then we apply the corresponding form of the universal approximation theorem to find an approximator which we will show will also be an interpolator.
	
	By the lemma \ref{h_continuous} we know that $H_i : \K \rightarrow \R^{q}$ (defined in \cref{eq:vector_distance}) are all continuous,  hence we can proceed to define the following vector valued function $H : \K \rightarrow \R^{q}$
	\begin{align}\label{def:H}
		H(x) = (H_1(x), H_2(x), \hdots, H_{q}(x)) ,
	\end{align}
	which must be continuous. Note that
	\begin{align*}
		p_{q}\circ H(x) =  \overline{f}(x) \quad x \in \M.
	\end{align*}
	As our activation function is a continuous non-polynomial, we can apply the universal approximation theorem \cite{pinkus_1999} on the function $H$. This guarantees us a single layer neural network $\Psi : \K \rightarrow \R^{q}$ such that
$
		\sup_{x \in \K} \|H(x) - \Psi(x)\| < \epsilon / 2 .
$
	We will show that
	\begin{align}
		p_{q}\circ \Psi(x) =  \overline{f}(x) \quad \forall x \in \M_\epsilon\cap\K.  \label{eq:result} 
	\end{align}
	Observe that on the sets $\M_\epsilon$ the function $H$ is of the form
$
		H(x) = \lambda*e_{\overline{f}(x)} 
$
	where $\lambda \in \R,  \lambda>\epsilon$ and $e_k\in \R^{q}$ is a k'th unit vector. Therefore,
	$
		\Psi(x) = (\psi_1(x), \psi_2(x), \hdots, \psi_{q}(x)) 
	$
	such that
	\begin{align*}
		\psi_i(x)  < \epsilon/2 \quad \text{if } i \neq  \overline{f}(x), \quad \psi_i(x)  > \epsilon/2 \quad \text{if } i =  \overline{f}(x). 
	\end{align*}
	The result \eqref{eq:result} follows immediately from this. This proves part \eqref{result:shallow_network}.
	
	For the \eqref{result:deep_network} we recall Theorem \ref{thrm:Lyons}.
	As our activation function was is non-polynomial, therefore it must also be non-affine, it satisfies all the conditions of Theorem \ref{thrm:Lyons} and the rest proceeds as in the shallow network case.
\end{proof}

\begin{remark}
	There are slightly stronger versions of this theorem. If the activation function is only continuous and non-polynomial, then there exists a shallow neural network that interpolates $f$ on $\M$. On the other hand, if the activation function is nonaffine continuous that is continuously differentiable at at least one point, with nonzero derivative at that point, then there exists a deep neural network with finite with that interpolates $f$ on $\M$. 
\end{remark}

An interesting note here is that one can notice that the function $H$ is in fact 1-Lipschitz, so the proof also shows that there exists a neural network that is stable in the Lipschitz framework. The caveat however is that in practice, the loss function is minimising the difference between $ \Psi$ and $\overline{f}$, not $p_{q}\circ \Psi$ with $\overline{f}$ which means that the algorithms usually do not converge at $H$.

\begin{proposition}\label{Lipschitz1}
	For the norm $\|\cdot\|_p$ where $1 \leq p \leq \infty$, the function $H:\R^d \rightarrow \R^{q}$ has Lipschitz constant 1.
\end{proposition}
\begin{proof}
	We want to show that 
	$
		\| H(x) - H(y)\|_p \leq \| x - y\|_p.
	$
	Recall that H is defined as the vector that consists of $H_i$ \cref{def:H}. From the \cref{eq:vector_distance} we see that $H(x)$ will have elements equal to 0, unless the index $i$ is equal to $\overline{f}(x)$. 
	
	Given this, we can distinguish two cases.
	  
	\textit{Case 1.} $\overline{f}(x) = \overline{f}(y)$
	We know that there is a sequence $\{z_i\}_{i=1}^\infty$ such that
	\begin{align}\label{lipschitz}
		\|x - z_i \|_p \rightarrow \h(x), \quad \text{where } \overline{f}(z_i)\neq \overline{f}(x),
	\end{align}
	and also
	$
		\|x - z_i \|_p\geq \h(x). 
	$
	Since $x,y$ have the same label, we obtain from \eqref{lipschitz} that 
	\begin{align*}
		\|H(x) - H(y)\|_p = |\h(x) - \h(y)| & \leq | \|x - z_i \|_p - \|y - z_i \|_p | \quad & \forall i \in \mathbb{Z}^{+}, \\
		| \|x - z_i \|_p - \|y - z_i \|_p |                                 & \leq \| x - y\|_p \quad                        & \forall i \in \mathbb{Z}^{+}. 
	\end{align*}
	  
	\textit{Case 2.} $\overline{f}(x) \neq \overline{f}(y)$
	In this case let us look at the line segment 
	\begin{align*}
		\mathcal{L} = \{tx + (1-t)y : t \in [0,1]\} ,
	\end{align*}
	and consider the following two points $w_1, w_2$ 
	\begin{align}
		w_1 & = t_1 x + (1-t_1)y \quad t_1  = \inf\{t : \overline{f}(tx + (1-t)y)\neq \overline{f}(x) \}, \label{infw} \\
		w_2 & = t_2 x + (1-t_2)y \quad t_2  = \sup\{t : \overline{f}(tx + (1-t)y)\neq \overline{f}(y) \}. \label{supw} 
	\end{align}
	Clearly $w_1 \leq w_2$, because otherwise $w_2 < \frac{w_1 + w_2}{2}< w_1$ and by the definitions \eqref{infw} \eqref{supw}
	\begin{align*}
		\overline{f}(\frac{w_1 + w_2}{2}) = \overline{f}(x)  \quad \text{as } \frac{w_1 + w_2}{2} < w_1, \\
		\overline{f}(\frac{w_1 + w_2}{2}) = \overline{f}(y)  \quad \text{as } \frac{w_1 + w_2}{2} > w_2. 
	\end{align*}
	This is a contradiction with $\overline{f}(x) \neq \overline{f}(y)$. Therefore $w_1 \leq w_2$ and hence 
	\begin{align*}
		\|H(x) - H(y)\|_p &= (|\h(x)|^p + |\h(y)|^p)^{1/p}  \leq ( |\|x - w_1 \|_p|^p + |\|y- w_2 \|_p|^p)^{1/p} \\
		& \leq \|x - w_1 \|_p + \|y- w_2 \|_p \leq \| x - y\|_p .
	\end{align*}
\end{proof}

Note that we could have also proven the theorem using Urysohn's lemma,  and we would obtain the same result. Using Urysohn's lemma we would construct a continuous function $H^* : \K \rightarrow \R^{q}$ such that
$
	p_{q}\circ H^*(x) = f(x),  
$	 
for all $x \in \M_\epsilon\cap\K.$
This would be done by applying Urysohn's lemma for indicator functions $\mathbbm{1}_i : \K \rightarrow \{0,1\}$ for each label $i \in \overline{\Y}$
\begin{align*}
	\mathbbm{1}_i(x) = 
	\begin{cases}
		1 \quad \text{if }f(x) = i, \\
		0 \quad \text{if }f(x) \neq i.
	\end{cases}	
\end{align*}
on disjoint subsets of $\M_{\epsilon}$, call this function obtained from Urysohn's lemma $U_i : \K \rightarrow [0,1]$. Then the final function $H^*$ would simply just be $H^*(x) = (U_1(x), U_2(x), \hdots, U_{q}(x))$.

The drawback here is that this function does not necessarily have a bounded Lipschitz constant. In the following examples we will illustrate that there are certain cases where the two functions $H$ and $H^*$ have different Lipschitz constants, yet their class stability is the same.
\begin{example}
Consider the classification function $f_l :[0,2] \rightarrow \{0,1\}$ where 
	\begin{align*}
		f_l = 
		\begin{cases}
			0 \quad \text{if }x<1, \\
			1 \quad \text{if }x \geq 1.
		\end{cases}
	\end{align*}
The $\M_{\epsilon}$ set for $\epsilon < 1$ here would therefore be the set $[0,1-\epsilon)\cup(1+\epsilon, 2]$. As we have shown in \cref{Lipschitz1}, the function $H$ will always have a Lipschitz constant of 1. However, the function $H^*$ will satisfy
\begin{align*}
	H^*(x) = 
	\begin{cases}
		(1,0) \quad \text{if }x<1-\epsilon, \\
		(0,1) \quad \text{if }x>1+\epsilon.  
	\end{cases}
\end{align*}
This means that we have a lower bound on the Lipschitz constant by $$L \geq \frac{\|(1,-1)\|_p}{2\epsilon}.$$ 
As this expression diverges as $\epsilon \rightarrow 0$, we see that the Lipschitz constant diverges as well. However, for both functions we have 
	\begin{align*}
		p_{q}\circ H(x) = p_{q}\circ H^*(x) = f_l(x) \quad \forall x \in \M_{\epsilon}.
	\end{align*}
Thus, $p_{q}\circ H$ and $p_{q}\circ H^*$ have the same class stability.
\end{example}

\section{Stability revised}
One relevant question one might have when talking about the class stability is how that relates to measure theory. In fact, if we were to look at the class stability from that point of view, one might argue that the functions mentioned in section \ref{example}, function $f_3$ might be considered the most stable and $f_1, f_2$ equally stable since the unstable points have measure 0.  We can define the class stability in the following sense to keep consistency.

\begin{definition}[Measure theoretic distance to the decision boundary]
	For a extension of a classification function $\overline{f} : \R^d \rightarrow \overline{\Y}$ and a real number $p \geq 1$, we define $\h: \R^d \rightarrow \R^+$ the \textit{$L^p$-distance to the decision boundary} as
	\begin{align*}
		\h(x) = \inf \{ r : \int_{\B^p_r(x)} \mathbbm{1}_{\bar{f}(z) = \bar{f}(x)}\, d\mu \neq\int_{\B^p_r(x)}\, d\mu, r\in[0,\infty)  \}. 
	\end{align*}
	Here $\mu$ denotes the Lebesgue measure and $\B^p_r(x)$ the unit closed ball with p-norm. One unfortunate thing for this definition is that the function is no longer continuous as can be seen by looking at the following function $f_2$ at the point $1/2$. The stability of that point is $0$, whereas now its neighbourhood has a non-zero stability as $1/2$ is an isolated point with a different label.  Fortunately we can show that the stability remains measurable if $f$ itself was measurable!
\end{definition}

\begin{lemma}[Measurability of stability]
	Let $f: \M \rightarrow \Y$ be a measurable classification function. Then the measure theoretic distance to the decision boundary $\h$ is measurable.
\end{lemma}
\begin{proof}
	Since $f$ is measurable, we know by Lusin's theorem that for any $\epsilon >0$ there exists a closed $F \subset \M$ such that 
	\begin{align*}
		|\M - F| < \epsilon, \quad f \text{ is continuous on } F .
	\end{align*}
	Fix an $\epsilon >0$ and the corresponding $\F\subset\M$. We will show that $\h$ is continuous on $F$.  By the continuity of $f$ on $F$ we have that for any sequence $x_i \rightarrow x \in F$, $f(x_i) \rightarrow f(x).$
	Since $f$ is a classification function, that means that eventually all $x_i$ have to have the same label as $x$ so without loss of generality,  let 
	$
		f(x_i) = f(x) \quad \forall i\in\mathbb{N}.
	$
	Define $R := \h(x)$, then by the triangle inequality we know that 
	\begin{align*}
		R - \|x-x_i \|^p \leq \h(x_i) \leq R + \| x - x_i\|^p \quad \forall i\in\mathbb{N}.
	\end{align*}
	Thus as $x_i \rightarrow x$, we obtain 
	$
		\h(x_i) \rightarrow \h(x),
	$
	for any sequence $\{x_i\} $ in $F$ which proves that $\h$ is continuous on F.
\end{proof}

For the rest of the document, we will always assume $f$ to be measurable and we will use $\h$ to refer to the measure theoretic distance to the decision boundary.
\section{Proof of \cref{exist_stable}}

We are now set to prove our next main result \cref{exist_stable}. To prove this theorem we will first show the following theorem.

\begin{proposition}\label{thm:1}
	Let $f: \M \rightarrow \Y$ be a classification function.  Then, for any set $\{(x_i, f(x_i))\}_{i=1}^k$  such that  $\h(x_i) > 0$ for all $i=1,\hdots, k$ (the distance to the decision boundary \cref{def:dist} is non-zero) and $\epsilon_1, \epsilon_2 >0$, there exists a continuous function $g: \M \rightarrow \R$ such that the class stability \cref{def:class_stability} satisfies
	\begin{align}\label{thm_stability}
		\eS^p_{\M}(\lfloor g\rceil) \geq \eS^p_{\M}(\overline{f})  - \epsilon_1
	\end{align}
	and the functions agree on the set
	\begin{align}\label{thm_accuracy}
		f(x_i) = g(x_i) \quad i = 1,\hdots ,k,
	\end{align}
	and 
	\begin{align}\label{thm_measure}
		\mu(R) < \epsilon_2, \quad R := \{x \, \vert \, f(x) \neq g(x), x \in \M\}, 
	\end{align}
	where $\mu$ denotes the Lebesgue measure and $\lfloor \cdot \rceil$ is the function that rounds to the nearest integer. 
\end{proposition}
	Note that the class stability of $\lfloor g \rceil$ is well defined as it is a discrete function defined on a compact set $\M$.
\begin{proof}[Proof of \cref{thm:1}]
	We define the following disjoint sets, based on the distance to the decision boundary function $\h$ \cref{def:dist}: For $\xi > 0$, let 
	\begin{align*}
		S_{\xi} &:= \{x \, \vert \, \h(x) \geq \xi , x \in \M\}, \quad U_{\xi} := \{x \, \vert \, \h(x) < \xi , x\in \M\}, \\
		 & \qquad \qquad \qquad U:= \{x | \h(x) = 0\ , x \in \M\}.
	\end{align*}
	First, notice that for any $\xi_1 < \xi_2$ we have $U_{\xi_1} \subset U_{\xi_2}$ and that for any $\eta >0 $ the following holds true 
	\begin{align}\label{U_converge}
		\bigcap\limits_{\xi < \eta} U_{\xi} = U.
	\end{align}
	Since $\h$ is measurable and we can write $U = \{x \, \vert \, \h(x) \leq 0\}$ as $\h$ is non-negative, we know that the set $U$ is measurable. In fact, by the same reasoning, all three sets are.  
	
	Consider the closure $\overline{S_\xi}$ of the set $S_\xi$, and the adjusted sets $U'_{\xi} = U_{\xi} - \overline{S_\xi}$ and $U^0_\xi = U - \overline{S_\xi}$.  As $\overline{S_\xi}$ is closed, it must be measurable and also the difference of two measurable sets is measurable, thus $\overline{S_\xi}, U'_\xi, U^0_\xi$ are all measurable.  
	
	\textbf{Claim 1}: $\mu(U \cap \overline{S_\xi}) = 0$.  To show the claim, we will start by considering the collection $\{B^p_{\xi/2}(x) \, \vert \, x \in S_\xi\}$ of open balls or radius $\xi$ in the p-norm, and noting that it is an open cover of $\overline{S_\xi}$. Therefore, since $\overline{S_\xi} \subset \mathcal{M}$, which is bounded, and since $\overline{S_\xi}$ is closed, there must exist a finite subcover, in particular there must exist a finite subset $S^* \subset S_\xi$ such that $\overline{S_\xi} \subset \bigcup_{x \in S^*} B^p_{\xi/2}(x)$. Now, suppose that $\mu(U \cap \overline{S_\xi}) > 0$, then we would neccesarily have 
	\begin{equation}
		\begin{split}
		\mu(U \cap (\bigcup_{x \in S^*} B^p_{\xi/2}(x))) > 0, \text{ hence }
		\mu(\bigcup_{x \in S^*} (U \cap B^p_{\xi/2}(x))) > 0.
		\end{split}
	\end{equation}
	By subadditivity (as $S^*$ is finite), there must exist a point $x_0$ such that $\mu(U \cap B^p_{\xi/2}(x_0)) > 0$. Recall that $x_0\in S_\xi$ means $\h(x_0) \geq \xi$ which implies  
	\begin{align}\label{eq:recall}
		\inf \{ r\in[0,\infty)  \, \vert \, \int_{\B^p_r(x_0)} \mathbbm{1}_{\bar{f}(z) = \bar{f}(x_0)}\, d\mu \neq\int_{\B^p_r(x_0)}\, d\mu \} \geq \xi.
	\end{align}
	Thus, the function $\overline{f}$ is constant on $B^p_{\xi/2}(x_0)$ almost everywhere and any point $z$ of the set
	\begin{align}\label{def:L}
		L_{x_0, \xi/2} : = \{z \, \vert \, z \in B^p_{\xi/2}(x_0), \, \overline{f}(z) = \overline{f}(x_0)\} 
	\end{align}
	satisfies $\h(z)\geq \xi/2$ as $x_0$ satisfies $\h(x_0) \geq \xi$. This means that $\mu(U \cap L_{x_0, \xi/2} ) = 0$ as all $z' \in U$ have $\h(z') = 0$ . Finally, from the fact that $\overline{f}$ is constant on $B^p_{\xi/2}(x_0)$ almost everywhere, we must have $\mu(B^p_{\xi/2}(x_0) - L_{x_0, \xi/2}) = 0$, which means that we cannot have $\mu(U \cap B^p_{\xi/2}(x_0)) > 0$, giving us the required contradiction and we have shown \textit{Claim 1}.
	
	\textbf{Claim 2}: \textit{$\overline{f}$ is continuous on $S_\xi$ and there exists a unique continuous extension of $\overline{f}$ to $\overline{S_\xi}$.} We start by showing that $\overline{f}$ is continuous on $S_\xi$. For any $x_0 \in S_\xi$ consider the neighbourhood $B^p_{\xi/2}(x_0)$ as before and recall that $\overline{f}$ is constant on this ball almost everywhere, with the constant being $\overline{f}(x_0)$. Suppose now that there is a $z \in S_\xi \cap B^p_{\xi/2}(x_0)$ such that $\overline{f}(x_0)\neq \overline{f}(z)$. As $z \in S_\xi$ (recall \eqref{eq:recall}), we must also have that $\overline{f}$ constant on $B^p_{\xi/2}(z)$ almost everywhere, with the constant being $\overline{f}(z)$. However, as $B^p_{\xi/2}(x_0)$ and $B^p_{\xi/2}(z)$ intersect we obtain our contradiction.  The second part of this claim follows a similar argument. Let $x^*$ be a limit point of $S_\xi$. Consider the set $B^p_{\xi/2}(x^*) \cap S_\xi$. By arguing as in the first part of the proof of the claim, no two points in this set can have different labels. Thus, this means that any sequence $x_i \rightarrow x^*\text{ as } i \rightarrow \infty$ with $x_i \in S_\xi$ we have $x_i \in B^p_{\xi/2}(x^*) \cap S_\xi$ for all large $i$, and thus all the labels will eventually have to be the same. Therefore, there is a unique way of defining the extension of $\overline{f}$ to $\overline{S_\xi}$, which proves \textit{Claim 2}. We will call this unique extension 
	\begin{equation}\label{f^*}
		\overline{f^*}: \overline{S_\xi} \rightarrow \overline{\Y}.
	\end{equation}
	
	\textbf{Claim 3}: \textit{Consider any $x_0 \in S_\xi$, and define $a = \h(x_0) - \xi$. We claim that $B^p_{a}(x_0) \subset \overline{S_\xi}$.} We first show that $\h \geq \xi$ on $B^p_{a}(x_0)$ almost everywhere for any fixed $x_0 \in S_\xi$. As before, it suffices to only consider the points $z\in B^p_{a}(x_0)$ such that $\overline{f}(z) = \overline{f}(x_0)$, as $\overline{f}$ is constant almost everywhere on this set. Suppose there exists $z \in L_{x_0,a}$ (as defined in \cref{def:L}) such that $\h(z) < \xi$. The ball centred at $x_0$ with a radius $\|x_0 - z\|_p + \h(z)$ has to contain the ball centred at $z$ with a radius of $\h(z)$. Thus, by the definition of the distance to the decision boundary, we must have $\h(x_0) <  \|x_0 - z\|_p + \h(z)< a + \xi = \h(x_0)$, which gives the contradiction. Therefore, $\h \geq \xi$ on $B^p_{a}(x_0)$ almost everywhere and hence
	\begin{align}\label{L_in_S} 
	L_{x_0,a} \subset S_\xi. 
	\end{align}
Now consider any $x\in B^p_a(x_0)$. Since the ball is open, there exists a $\delta_0 > 0$, such that $B^p_\delta(x) \subset B^p_a(x_0)$ for all $\delta < \delta_0$. Moreover, as $\mu(B^p_\delta(x))>0$ for any $\delta > 0$, there must be a sequence $\{x_i\}^\infty_{i=1} \subset L_{x_0, a}$ such that $x_i \rightarrow x$ as $i \rightarrow \infty$, as $L_{x_0, a} \subset B^p_a(x_0)$ and $\mu( B^p_a(x_0) - L_{x_0, a} ) = 0$. This means that $x \in \overline{L_{x_0, a}}$ the closure of $L_{x_0, a}$ and from \cref{L_in_S} we obtain $x \in \overline{S_\xi}$ for all $x\in B^p_a(x_0)$.  Therefore $B^p_a(x_0) \subset \overline{S_\xi}$ which proves \textit{Claim 3}.
	
	Next we apply Lusin's Theorem for the function $\overline{f}$ on the set $U^0_\xi$ and obtain, for any $\alpha > 0$, a closed set $U^{\alpha}_\xi \subset U^0_\xi$ such that 
	\begin{align}\label{lusin_bound}
		\mu(U^0_\xi-U^{\alpha}_\xi) < \alpha, \quad \overline{f} \text{ is continuous on } U^{\alpha}_\xi.
	\end{align}

We can now define $g_{\alpha, \xi}: \overline{S_{\xi}}\cup U^{\alpha}_\xi \rightarrow [a,b]$,  where $a := \min \{ \overline{\Y} \}$ and $b := \max \{ \overline{\Y} \}$, where 
	\begin{align*}
		g_{\alpha, \xi}(x) = 
		\begin{cases}
			 \overline{f^*}(x) \quad \text{if }x\in \overline{S_\xi}, \\
			\overline{f}(x) \quad \text{if }x\in U^{\alpha}_\xi.
		\end{cases}
	\end{align*}
	Finally, as both sets $\overline{S_\xi}$ and $U^{\alpha}_\xi$ are compact, since they are closed and subsets of $\M$ which is compact, we can apply Tietze's extension theorem. More precisely, we will use Tietze's extension theorem to extend the restriction of the function $g_{\alpha, \xi} : \overline{S_{\xi}}\cup U^{\alpha}_\xi \rightarrow [a, b]$, to a continuous function on the whole set $\M$. Then by Tietze's extension theorem we obtain a continuous function $g^*_{\alpha, \xi}: \M \rightarrow [a,b]$ such that 
	\begin{align*}
		g^*_{\alpha, \xi}(x) = g_{\alpha, \xi}(x) \quad x \in \overline{S_{\xi}}\cup U^{\alpha}_\xi.
	\end{align*}
	
	Having constructed the function, all we need to do is to check that the stability and accuracy satisfies properties (\ref{thm_stability})  (\ref{thm_accuracy}) and (\ref{thm_measure}) for some particular choices of $\alpha$ and $\xi$.
	Let us first estimate the loss in class stability for the rounded function $\lfloor g^*_{\alpha, \xi}\rceil$. For any fixed $\xi$ we can bound the stability by:
	\begin{align*}
		\eS^p_{\lfloor g^*_{\alpha, \xi} \rceil} = \int_\mathcal{M} h^p_{\lfloor g^*_{\alpha, \xi} \rceil}\, d\mu = \int_{\overline{S_{\xi}} \cup U'_{\xi}} h^p_{\lfloor g^*_{\alpha, \xi} \rceil}\, d\mu.
	\end{align*}
We know that $\overline{f^*}$ (defined in \cref{f^*}) and $g^*_{\alpha, \xi}$ agree on $\overline{S_{\xi}}$, hence $\lfloor g^*_{\alpha, \xi} \rceil$ agrees with $\overline{f^*}$ as well. From \textit{Claim 3} we know that for any point $x_0 \in S_\xi, \, B^p_{a}(x_0) \subset \overline{S_\xi}$, where $a = \h(x_0)-\xi$,  while from \textit{Claim 2} we know that $\overline{f^*}$ is continuous on $\overline{S_\xi}$, therefore $\overline{f^*}$ is constant on $B^p_{a}(x_0)$ as it is a discrete function. Thus we must have $h^p_{\lfloor g^*_{\alpha, \xi} \rceil}(x_0) \geq \h(x_0)-\xi$ for all $x_0 \in S_\xi$. This means that
	\begin{align*}
		\eS^p_{\lfloor g_{\alpha, \xi} \rceil} &= \int_{\overline{S_{\xi}} \cup U'_{\xi}} h^p_{\lfloor g_{\alpha, \xi} \rceil}\, d\mu \geq 
		\int_{S_{\xi} \cup U'_{\xi}} h^p_{\lfloor g_{\alpha, \xi} \rceil}\, d\mu \geq 
		\int_{S_{\xi} } \h -\xi \, d\mu \\
		& =\int_{\M - U_\xi} \h \, d\mu - \xi \mu(S_{\xi})  = \eS^p(f) - \int_{U_\xi}\h \, d\mu - \xi \mu(S_{\xi}) \\
		&> \eS^p(f) - \xi \mu(U_\xi) - \xi \mu(S_\xi) = \eS^p(f) - \xi \mu(\M).
	\end{align*}
The last inequality comes from the fact that $\h(x) < \xi$ for $x \in U_\xi$. By choosing $\xi \leq \frac{\epsilon_1}{\mu(\M)}$ we obtain \cref{thm_stability}. 

To ensure the (\ref{thm_accuracy}) we simply need to guarantee that the set $\{x_i\}_{i=1}^k$, from the statement of the proposition, satisfies $\{x_i\}_{i=1}^k \subset S_{\xi}$. This can be achieved by choosing $\xi < \min_{i=1,\hdots,k}\{\h(x_i)\}$.
	
Finally,  we observe that $R \subset U'_{\xi} - U^\alpha_{\xi}$, where we recall $R$ from \cref{thm_measure}. Therefore, we have
	\begin{equation}
	\begin{split}
			\mu(R) &\leq \mu(U'_{\xi} - U^\alpha_{\xi}) \leq \mu(U'_{\xi} - U^0_{\xi}) + \mu(U^0_{\xi} - U^\alpha_{\xi}) \\
			& < \mu(U'_\xi-U^0_\xi) + \alpha 
			 = \mu((U_\xi - \overline{S_\xi}) - (U-\overline{S_\xi})) + \alpha = \mu(U_\xi - U) + \alpha.
	\end{split}
	\end{equation}
	Thus, to establish \cref{thm_measure}, it suffices to show that $\mu(U_\xi) \to \mu(U)$ as $\xi \to 0$, and then by setting $\alpha = \epsilon_2/2$ we could choose a small enough $\xi$ to finally obtain (\ref{thm_measure}). Thankfully, this is true as we have shown that $U_{\xi}$ is decreasing in $\xi$ and since $U_{\xi} \subset \M$, we know that the measure $\mu(U_{\xi})\leq\mu(\M)$. Therefore, $\mu(U_{\xi})$ is bounded and because of \cref{U_converge} we can apply Theorem 3.26 from \cite{wheeden_zygmund} to obtain $\mu(U_\xi) \to \mu(U)$ as $\xi \to 0$.
\end{proof}

\begin{proof}[Proof of \cref{exist_stable}]
	Using \cref{thm:1} we construct a continuous function $g: \M \rightarrow \R$ that satisfies the conditions.  Next we construct a continuous function $G: \M \rightarrow \R^{q}$ such that
	\begin{align}
		 \eS^p_\M(p_{q}(G)) \geq \eS^p_\M(\overline{f}) - \epsilon_1,
	\end{align}
	we can interpolate on the set
	\begin{align}
		p_{q}(G) = f(x_i) \quad i = 1,\hdots,k \,,
	\end{align}
	and
	\begin{align}
		\mu(R) < \epsilon_2, \quad R := \{x \, \vert \, f(x) \neq p_{q}(G), x \in \M\}, 
	\end{align}
	where $\mu$ denotes the Lebesgue measure. 
	Recall from the proof of \cref{thm:1} that $g$ is constant on $\overline{S_\xi}\cup U^\alpha_\xi$ for $\xi>0$. Furthermore,  from the proof it is clear that any function that agrees with $g$ on the set $\overline{S_\xi}\cup U^\alpha_\xi$ will also have to satisfy all three conditions of the theorem. Therefore, it is enough to construct $G$ such that $p_{q}(G)$ agrees with $g$ on $\overline{S_\xi}\cup U^\alpha_\xi$. To construct the function $G$, consider the function $\omega: \R \rightarrow \R$ defined by
	\begin{align}
		\omega_i(x) = 
			\begin{cases}
				0 \quad &x \leq i-1, \\
				x-(i-1) \quad &i-1<x \leq i, \\
				(i+1)-x \quad &i<x\leq i+1, \\
				0 \quad  &i+1 \leq x.
			\end{cases}
	\end{align} 
Having this, we can simply define $G(x) = (\omega_{1}(g(x)), \dots, \omega_{q}(g(x)))$ which will be continuous as $\omega$ is continuous. Furthermore, it agrees with $g$ on $\overline{S_\xi}\cup U^\alpha_\xi$ and thus satisfies all three conditions of the theorem. We now just need to apply the universal approximation theorem on the function $G$ to obtain a neural network $\psi: \M \rightarrow \R^{q}$ that differs from $G$ in the uniform norm by less than $1/2$. This neural network will give the same labels on $\overline{S_\xi}\cup U^\alpha_\xi$ as $G$ and thus must satisfy all three conditions of the theorem, thereby completing the proof.
\end{proof}

\bibliographystyle{abbrv}
\bibliography{references}
\end{document}